\documentclass[11pt]{article}

\usepackage[T1]{fontenc}
\usepackage[utf8]{inputenc}
\usepackage{lmodern}
\usepackage{microtype}
\usepackage[margin=1in]{geometry}
\usepackage{amsmath,amssymb,mathtools}
\usepackage{booktabs,tabularx,multirow}
\usepackage{enumitem}
\usepackage{xcolor}
\usepackage{graphicx}
\usepackage{float}
\usepackage{tikz}
\usetikzlibrary{arrows.meta,backgrounds,fit,positioning}
\usepackage[numbers,sort&compress]{natbib}
\usepackage[hidelinks]{hyperref}
\usepackage[nameinlink,noabbrev]{cleveref}
\hypersetup{
  pdftitle={Governed Persistent Memory: Source-Bound State Semantics and Fail-Closed Release for Long-Horizon Agents},
  pdfauthor={Guodong Xu},
  pdfsubject={Auditable state semantics and bounded end-to-end evaluation for persistent agent memory},
  pdfkeywords={persistent memory, AI agents, provenance, bitemporal state, fail-closed release, evaluation}
}
\newcolumntype{P}[1]{>{\raggedright\arraybackslash}p{#1}}
\newcolumntype{R}[1]{>{\raggedleft\arraybackslash}p{#1}}
\newcolumntype{Y}{>{\raggedright\arraybackslash}X}

\definecolor{govblue}{HTML}{2E73B8}
\definecolor{govfill}{HTML}{EAF3FA}
\definecolor{rankgray}{HTML}{66758A}
\definecolor{rankfill}{HTML}{F2F5F8}
\definecolor{releaseorange}{HTML}{B96C10}
\definecolor{releasefill}{HTML}{FFF5E9}
\definecolor{boundaryred}{HTML}{B52222}
\newcommand{\sys}{GPM}
\newcommand{\code}[1]{\texttt{#1}}

\title{Governed Persistent Memory:\\
Source-Bound State Semantics and Fail-Closed Release for Long-Horizon Agents}
\author{Guodong Xu\\
\small Qingdao Guodongxiansheng Network Technology Co., Ltd.\\
\small (Gu\v{o}d\`ong Xi\=ansheng)\\
\small \texttt{kzkz137806@gmail.com}}
\date{August 2026}

\begin{document}
\maketitle

\begin{abstract}
Long-term agent memory is usually treated as select--store--retrieve, but retrieval does not decide whether contradictory, superseded, retracted, deleted, or stale records may support an outgoing claim.  We introduce \emph{Governed Persistent Memory} (\sys), an auditable bitemporal state-transition model with source-bound admission, derived lifecycle state, current public barriers, and fail-closed structured release.  Five executable clauses cover ledger integrity, source binding, conflict isolation, non-revival after retraction or deletion, and exact claim closure over a fresh view at one verified head.

On a prespecified hash-frozen 3,600-case GPM-ReleaseBench, \sys{} matches all complete outcomes; the strongest of three intentionally simple complete policies matches 1,800/3,600 and makes unmatched releases on 50\% of violation cases.  A separate sealed end-to-end service evaluation exercises real ingestion and release across eight query families.  In its publicly disclosed V3 arm, the governed lane is correct on 2,400/2,400 clusters versus 600/2,400 for ungoverned local Qwen2.5-7B; it repairs all 1,800 baseline failures with no regression (one-sided 95\% lower bounds 99.875\% and 99.834\%).  A later V5 reseal over Chinese- and English-command arms, with generation-date pinning and no post-freeze reducer amendment, again obtains 2,400/2,400 per arm.  A production-code-independent finite model explores 331,776 semantic and 1,990,656 query states without a full-contract counterexample, and a 100,000-trace three-engine differential yields zero mismatches.

These are bounded contract and implementation results, not open-world model accuracy or evidence of world truth.  Governed answers in the sealed service evaluation are deterministic service outputs; the 7B result is the ungoverned comparison, not a claim that a language model itself became perfectly accurate.
\end{abstract}

\section{Introduction}

Long-horizon agents must preserve facts, preferences, commitments, and task state across interactions.  Retrieval-augmented generation, generative agents, external memory, and memory operating systems extend a model beyond its immediate context window \citep{lewis2020rag,park2023generative,packer2023memgpt,wang2023longmem,chhikara2025mem0,li2025memos}.  Yet retrieving a relevant passage and being entitled to assert a current fact are different operations.

Consider four common changes.  A user moves from one city to another.  Two sources disagree about a single-valued attribute.  A previously asserted fact is later retracted.  A user deletes their history, but a stale snapshot or derived index can still surface it.  Relevance ranking can return a high-scoring candidate in every case; relevance alone does not establish public assertability.  Persistence also changes the time scale of failure: one extraction error can be recalled, summarized, and written back until it acquires the appearance of stable history.  HaluMem localizes such propagation across extraction, update, and question-answering stages \citep{chen2025halumem}; LongMemEval, LoCoMo, and MemoryAgentBench measure complementary aspects of long-term interactive memory \citep{maharana2024locomo,wu2025longmemeval,hu2025memoryagentbench}.

We ask a narrower systems question: \emph{what state semantics must hold before persistent memory may support a public claim?}  Our answer, \sys, separates external source acquisition, authenticated episode registration, optional claim admission, append-only history, public projection, candidate ranking, untrusted claim proposal, and a local structured-release decision.  The contribution is not a new vector index.  It is a set of executable state obligations and a release boundary that remains meaningful when records conflict, arrive late, are retracted, or must become publicly inaccessible.

Several 2026 systems sharpen this boundary.  StateFuse preserves contradictions across replicated operation sets and resolves them at projection time \citep{volkov2026statefuse}; TOKI defines a bitemporal operator algebra for contradictory histories \citep{wang2026toki}; MemIR separates evidence, assertion, and decision roles to prevent provenance-role collapse \citep{jin2026memir}; and MemoRepair withdraws and selectively rebuilds provenance-linked descendants after invalidation \citep{zhao2026memorepair}.  \sys{} addresses a narrower end-to-end state contract: source admission into a bitemporal public state, current-policy non-revival across historical reads, and exact closure of an outgoing structured claim set over a fresh view under one stable verified head.  It does not provide replica convergence, semantic tombstones, arbitrary descendant repair, or a guarantee for free-form generation.

The paper makes six contributions:

\begin{enumerate}[leftmargin=*,itemsep=2pt]
  \item a bitemporal, source-bound event model with derived rather than caller-assigned lifecycle states;
  \item five executable contract clauses linking storage integrity, source binding, conflict isolation, retraction and deletion barriers, and structured release;
  \item GPM-ReleaseBench v1, an internally designed contract-conformance suite with frozen development and evaluator-hidden counterfactual cases plus matched complete policies;
  \item a sealed end-to-end evaluation through the real ingestion and release services, with eight governed query families, an ungoverned local 7B comparison, and a bilingual command-surface reseal;
  \item bounded exhaustive checking and 100,000-trace differential evidence across strict, incremental, and segmented implementations; and
  \item clean-task utility controls and an explicit failed language-judge gate that prevent state-level guarantees from being rewritten as open-world answer correctness.
\end{enumerate}

\section{From Retrieval to Governed State}

\subsection{Retrieval is not public eligibility}

Let $M(u,t)$ be the ungoverned set of a user's stored memory items visible at valid-time point $t$, and let $k$ be a fixed top-$k$ truncation depth.  A retriever returns ranked candidates
\begin{equation}
R(q,u,t)=\operatorname{rank}_{k}\{s(q,m_j):m_j\in M(u,t)\},
\end{equation}
where $q$ is a query, $u$ a user, $t$ a valid-time point, and $s$ any lexical, vector, graph, or learned score, and $\operatorname{rank}_{k}$ orders items by score and keeps the $k$ highest.  The score estimates relevance; it does not decide whether $m_j$ has admissible provenance, is superseded, is in unresolved conflict, is retracted, or is hidden by a deletion barrier.  We therefore define an assertable-fact projection $V_{\mathrm{pub}}$ and a testimony-only episode projection $E_{\mathrm{pub}}$ before ranking:
\begin{equation}
R_{\sys}(q,u,t,\tau;h_c)=\operatorname{rank}_{k}\{s(q,z):z\in V_{\mathrm{pub}}(u,t,\tau;h_c)\cup E_{\mathrm{pub}}(u,t,\tau;h_c)\}.
\end{equation}
Here $h_c$ is the engine's current verified ledger head, defined in the formal model below.  The projections determine candidate eligibility.  Ranking only orders projected candidates and provides no end-to-end correctness guarantee.  In particular, an episode in $E_{\mathrm{pub}}$ may be retrieved as testimony but cannot satisfy the structured release contract.

\subsection{Failure layers}

The architecture separates four failure layers.  At the \emph{storage layer}, events may be altered, truncated, duplicated, or reordered.  At the \emph{state layer}, unbound or caller-inferred claims, invalid supersession, unresolved conflict, or deletion revival can enter a public view.  At the \emph{retrieval layer}, relevant evidence may be missed or irrelevant evidence returned.  At the \emph{language layer}, a generator can negate, merge, overgeneralize, or fabricate even when evidence is present.  The first two layers admit deterministic contract checks; the latter two require empirical evaluation and, for free-form language, stronger semantic checking than the present system provides.

\section{Formal Model}

\subsection{Event ledger and trusted commitments}

An event is
\begin{equation}
e_i=(i,\tau_i,t_i,\mathrm{type}_i,\mathrm{payload}_i,h_{i-1},h_i),
\qquad
h_i=H(\operatorname{canon}(e_i\setminus\{h_i\})).
\end{equation}
Here $i$ is a contiguous sequence number, $\tau_i$ is transaction time, $t_i$ is an optional valid-time field, and $h_i$ is a canonical hash chained to $h_{i-1}$.  A valid ledger requires nondecreasing transaction time, $\tau_i\geq\tau_{i-1}$.  Transaction time is normally assigned by the engine clock; an explicitly trusted ingestion or reproducibility interface may inject a historical value only when the append preserves this order.  A decreasing injected value is rejected before the event is committed.

The single-file engines keep event count and chain head in a separately protected sidecar commitment.  The segmented engine commits event count, chain head, segment metadata, and for every user-index shard both a record count and rolling hash head.  A missing, modified, injected, or truncated index reference causes the affected public read to fail closed; full audit validates every declared or present shard.  Indexes are derived from authoritative segments and can be rebuilt, but are not trusted merely because they are derivable.

The state is $\Sigma=(L,C,P,B,I)$: authoritative ledger $L$, materialized claims $C$, provenance witnesses $P$, public barriers $B$, and rebuildable indexes $I$.  Events are applied through a partial transition $\delta(\Sigma,e)=\Sigma'$.  If the sequence, predecessor, hash, commitment, or public-eligibility precondition fails, public reads stop or the affected claim remains outside the public projection.

\subsection{Claims, provenance, and lifecycle}

A claim is
\begin{equation}
c=(u,f,x,a,v,[t_s,t_e),\tau,p,r,\sigma),
\end{equation}
with user $u$, fact identifier $f$, entity $x$, attribute $a$, value $v$, valid interval $[t_s,t_e)$, transaction time $\tau$, provenance witness $p$, optional superseded fact $r$, and descriptive support label $\sigma$.  The executable checks act through $p$, not through $\sigma$.  A direct episode witness binds the source event identifier and hash and requires a non-empty quote contained in the source.  A digestion-ledger witness additionally binds claim, ledger, source unit, content, quote, extraction method, extractor version, extraction time, and byte span.  Quote containment establishes string origin and location only; it is not a semantic-entailment test.  A caller-marked inference is quarantined and cannot become a public fact through this admission path, but a semantically unsupported extraction can still pass if its quoted bytes exist.  Semantic faithfulness therefore remains outside I2.

Lifecycle state is derived from events rather than stored as a freely mutable label.  The implemented transition vocabulary is summarized in \cref{tab:transitions}.

\begin{table}[t]
\centering
\small
\caption{Implemented event transitions.  ``Public effect'' is evaluated at the requested valid and transaction snapshot, subject to current public barriers.}
\label{tab:transitions}
\begin{tabularx}{\textwidth}{@{}P{0.22\textwidth}P{0.40\textwidth}Y@{}}
\toprule
Event & Preconditions and state update & Public effect \\
\midrule
\code{episode.add} & Append a source episode for one user; bind system transaction time. & May appear as testimony-only retrieval material; never assertable under I5. \\
\code{fact.quarantine} & Record a failed governed claim admission against a bound episode. & Claim is not assertable; the governed input episode is suppressed publicly. \\
\code{fact.assert} & Require a direct source-bound witness; validate valid interval and any supersedes edge. & Eligible if active, nonconflicting, and unblocked. \\
\code{fact.retract} & Target an earlier fact owned by the same user. & Target fact and its bound episode are absent from current public views. \\
\code{user.delete} & Establish the latest public deletion barrier for the user. & Events at or before the barrier cannot reappear publicly. \\
\bottomrule
\end{tabularx}
\end{table}

\subsection{Conflict and supersession}

Attribute cardinality is a deployment policy.  The prototype treats unspecified keys as single-valued and includes common plural keys such as \code{tag/tags}, \code{label/labels}, and \code{interest/interests} in a built-in multi-value registry; deployment additions extend rather than replace this registry.  Normalization $N$ performs Unicode NFKC normalization, trimming, and case folding only.  It performs no unit conversion or ontology reasoning.

Two active claims conflict, $c_i\bowtie c_j$, when they have the same user, normalized entity, and single-valued attribute; overlapping valid intervals; unequal normalized values; and neither a retraction nor a valid supersession relation resolves them.  A supersedes edge is valid only when its target is earlier and has the same user, normalized entity, and attribute.  An invalid edge neither suppresses its target nor passes full audit.  Unresolved sides are isolated from the public projection.

\subsection{Public and audit views}

Because transaction time is nondecreasing, let $L_{\leq\tau_q}(h_c)$ be the longest sequence prefix whose events have transaction time no later than $\tau_q$ within the ledger authenticated by the current verified head $h_c$, and let $C_{\leq\tau_q}(h_c)$ be the claims materialized from that prefix.  The current public policy for user $u$ at $h_c$ is $\Pi_{\mathrm{pub}}(u;h_c)=(d_c,R_c)$, where $d_c$ is the sequence of the latest \code{user.delete} event (or zero) and $R_c$ contains fact identifiers retracted after that barrier in the full prefix through $h_c$.  Thus transaction time chooses the historical evidence prefix, while the current head supplies non-revival policy.  Define
\begin{align}
\mathcal{C}_q(u,t;h_c)=\{c\in C_{\leq\tau_q}(h_c):{}&c.u=u,\;
\operatorname{seq}(c)>d_c,\;t\in[c.t_s,c.t_e),\notag\\
&c.f\notin R_c,\;\operatorname{prov}_{\leq\tau_q}(c;h_c)\},\\
V_{\mathrm{pub}}(u,t,\tau_q;h_c)=\{c\in\mathcal{C}_q:{}&
\neg\operatorname{superseded}_{\mathcal{C}_q}(c)\land
\neg\operatorname{conflicted}_{\mathcal{C}_q}(c)\}.
\end{align}
Here $\operatorname{seq}(c)$ is the sequence of the assertion event.  The source-binding predicate is evaluated against the requested prefix after the current deletion barrier; consequently a post-deletion claim cannot revive a pre-deletion source episode.  Supersession and conflict are then derived over the remaining requested-prefix candidates.  A retraction or deletion written after $\tau_q$ can therefore suppress a public historical read, but an assertion written after $\tau_q$ cannot enter it.  The default public mode sets \code{transactionAt=null} and reads the current prefix.  A non-null $\tau_q$ explicitly requests a policy-filtered historical prefix; it is not a current-state answer mode, because a contradiction or superseding assertion written after $\tau_q$ is absent from that prefix.

An authorized audit view instead derives its deletion and retraction policy from $L_{\leq\tau_q}$ itself and can reconstruct ordinary as-of state.  In the implementation, \code{auditView} and \code{auditSearch} are separate methods, while public \code{view/search} and the public adapter reject audit flags.  This API separation prevents accidental routing but is not itself authentication or authorization.

The implementation also derives $E_{\mathrm{pub}}(u,t,\tau_q;h_c)$ from episode events that survive the current deletion, quarantine, and fact-retraction barriers.  An otherwise eligible episode can appear in this set without any admitted fact and is labeled \code{testimony-only}.  It can be ranked for recall, but the local release decision indexes only assertable facts in $V_{\mathrm{pub}}$.

\subsection{Executable obligations}

The implementation checks five obligations:

\begin{description}[leftmargin=2.4em,style=nextline,itemsep=2pt]
  \item[I1 --- Ledger integrity.] Relative to a trusted local commitment, sequence, predecessor, canonical event hash, segment inventory, and index-shard commitments agree.  Failure stops public service or enters an explicit recovery path.
  \item[I2 --- Syntactic source binding.] Every public fact has a valid, direct, same-user witness bound to source bytes and location metadata.  Caller-declared inferred facts are quarantined or rejected before public eligibility, and projection rechecks source-binding eligibility rather than trusting admission permanently.  I2 does not establish that the quoted text semantically entails the normalized claim.
  \item[I3 --- Conflict isolation.] No unresolved pair $c_i\bowtie c_j$ appears as simultaneously assertable public facts.
  \item[I4 --- Public non-revival.] Public reads apply the retraction and user-deletion policy at the current verified head even when $\tau_q$ requests an earlier prefix.  A covered fact and its bound episode cannot re-enter through historical time, stale engine indexes, or engine-maintained materialized projections.
  \item[I5 --- Structured release closure.] Every released structured answer claim exactly matches an assertable fact in the fresh public view and supplies a non-empty set of source fact identifiers, all of which match that claim.  The returned local decision record binds the canonical complete claim multiset, policy and normalizer versions, user, query, snapshots, and verified head.  Release is permitted only if the verified head remains $h_b$ while the decision is formed; a changed head fails closed without a record.
\end{description}

For a structured claim $a$ (here $a$ always denotes a structured claim, not the claim-tuple attribute field), let $S(a)$ be its supplied source fact identifiers and let $\operatorname{tri}(a)$ be its normalized entity--key--value triple.  Given a public view $V$, define
\begin{equation}
\begin{aligned}
\mathcal{M}(a,V,u)=\{c.f\mid {}& c\in V,\ c.u=u,\\
 & \operatorname{tri}(c)=\operatorname{tri}(a)\},\\
\operatorname{bound}(a,V,u)\iff {}& S(a)\neq\varnothing\land S(a)\subseteq\mathcal{M}(a,V,u).
\end{aligned}
\end{equation}
Let $\gamma(A)$ be the lexicographically sorted multiset of each claim's normalized entity--key--value triple together with the sorted unique identifiers in $S(a)$.  Its canonical representation is a typed, length-delimited encoding prefixed by the domain and schema tag \code{GPM-CLAIMS-v1}; field types and multiplicity are preserved rather than joined with an ambiguous delimiter.  Define
\begin{equation}
D_A=H(\operatorname{canon}(\gamma(A))).
\end{equation}
For user $u$, query $q$, requested snapshot $(t,\tau)$, policy version $\nu_P$, and normalizer version $\nu_N$, define the local record payload
\begin{equation}
\rho(d)=\langle d,u,H(q),D_A,t,\tau,h_b,\nu_P,\nu_N\rangle.
\end{equation}
The gate reads a verified head $h_b$, constructs $V_b=V_{\mathrm{pub}}(u,t,\tau;h_b)$ itself, evaluates every binding against $V_b$, and then reads the verified head again as $h_a$.  Its result is
\begin{equation}
\label{eq:release-decision}
\operatorname{release\_decision}(A,u,q,t,\tau)=
\begin{cases}
\bot_{\mathrm{fail}}, & h_b\neq h_a,\\
\rho(\textsc{release}),
 & h_b=h_a\land A\neq\varnothing\land \forall a\in A:\operatorname{bound}(a,V_b,u),\\
\rho(\textsc{abstain}), & h_b=h_a\land\text{otherwise.}
\end{cases}
\end{equation}
Here $\bot_{\mathrm{fail}}$ means that the prototype raises an error and returns no decision record.  The gate does not accept caller-fabricated ``assertable'' objects.  A verifier recomputes $D_A$, checks all context and version fields, and rejects a record whose ledger head is no longer current.  The query hash binds record context but does not affect the release predicate; neither it nor I2 proves that a structured claim answers the query, follows semantically from its quote, or survives a later free-text rewrite.  The prototype record is not signed or transferable: it is valid only inside the trusted local process boundary.  A cross-process or cross-host credential would additionally require authentication such as a MAC or signature, anti-replay freshness, expiry, and key management.  Equation~\eqref{eq:release-decision} is the operational expansion of I5 for well-formed inputs.

\paragraph{Contract consequence 1 (claim-bound non-revival).}
Assume I1--I5 hold under the stated threat model and the decision defined in \cref{eq:release-decision} returns a local record whose decision is \(\textsc{release}\).  Then every \(a\in A\) is exactly bound to one or more same-user assertable facts in the fresh \(V_{\mathrm{pub}}(u,t,\tau;h_b)\), $D_A$ binds the complete canonical claim multiset and its supplied fact identifiers, and no \(a\) is bound only to a fact or episode excluded by the current retraction or user-deletion policy at $h_b$.  If the verified head changes while the record is constructed, the prototype fails closed by raising an error; no decision record is returned.

\emph{Derivation.}  The result follows directly from the contract definitions rather than an inductive theorem over all reachable states.  Equation~\eqref{eq:release-decision} requires $h_b=h_a$, $A\neq\varnothing$, and $\operatorname{bound}(a,V_b,u)$ for every $a\in A$; $D_A$ commits to the canonical complete multiset.  The definition of $\operatorname{bound}$ gives exact normalized values and a non-empty subset of matching fact identifiers.  Membership in $V_b$ entails syntactic source-binding eligibility by I2 and exclusion of unresolved conflicts and current non-revival barriers by I3--I4.  I1 and the before/after head check bind those predicates to one verified ledger head; a changed head instead selects $\bot_{\mathrm{fail}}$.  This consequence is source-relative and applies only to explicit structured claims; it proves neither semantic entailment, query relevance, nor properties of later free text.

\section{System Design}

\subsection{Trust boundary and pipeline}

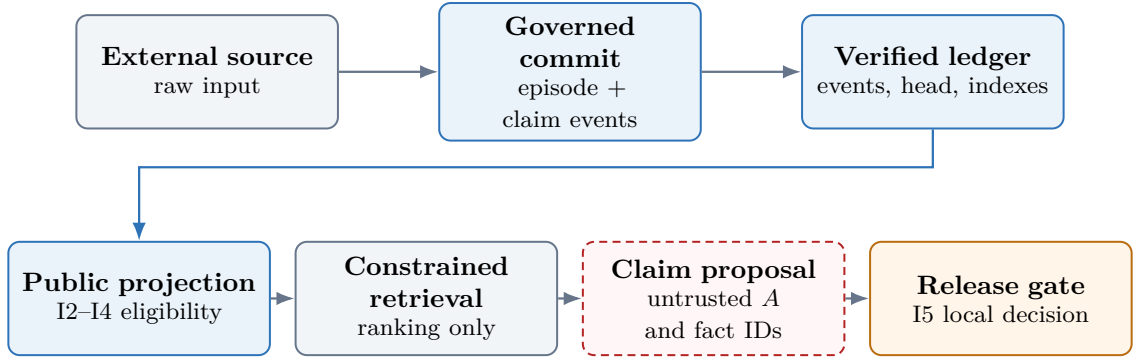
\begin{figure}[H]
\centering
\begin{tikzpicture}[
  >=Latex,
  box/.style={draw,rounded corners=1.5mm,text width=31mm,minimum height=15mm,align=center,line width=.75pt,font=\small,inner sep=1.8mm},
  flow/.style={->,line width=.85pt,draw=rankgray}
]
  \node[box,draw=rankgray,fill=rankfill] (source) at (-4.8,1.55) {\textbf{External source}\\[-.5mm]\footnotesize raw input};
  \node[box,draw=govblue,fill=govfill] (commit) at (0,1.55) {\textbf{Governed commit}\\[-.5mm]\footnotesize episode + claim events};
  \node[box,draw=govblue,fill=govfill] (ledger) at (4.8,1.55) {\textbf{Verified ledger}\\[-.5mm]\footnotesize events, head, indexes};

  \node[box,draw=govblue,fill=govfill] (projection) at (-5.7,-1.45) {\textbf{Public projection}\\[-.5mm]\footnotesize I2--I4 eligibility};
  \node[box,draw=rankgray,fill=rankfill] (retrieval) at (-1.9,-1.45) {\textbf{Constrained retrieval}\\[-.5mm]\footnotesize ranking only};
  \node[box,draw=boundaryred,densely dashed,fill=red!3] (proposal) at (1.9,-1.45) {\textbf{Claim proposal}\\[-.5mm]\footnotesize untrusted $A$ and fact IDs};
  \node[box,draw=releaseorange,fill=releasefill] (release) at (5.7,-1.45) {\textbf{Release gate}\\[-.5mm]\footnotesize I5 local decision};

  \draw[flow] (source) -- (commit);
  \draw[flow] (commit) -- (ledger);
  \draw[flow,draw=govblue] (ledger.south) -- ++(0,-5mm) -| (projection.north);
  \draw[flow] (projection) -- (retrieval);
  \draw[flow] (retrieval) -- (proposal);
  \draw[flow] (proposal) -- (release);
\end{tikzpicture}
\caption{Public-memory path and trust boundaries.  Raw acquisition is outside governed state; the authenticated ledger is the source of public projection.  I2 establishes syntactic source binding, not semantic entailment.  Retrieval ranks eligible candidates but neither admits claims nor authorizes release.  Lifecycle writes are listed in \cref{tab:transitions}, and the local-decision protocol is expanded in \cref{fig:release-protocol}.}
\label{fig:pipeline}
\end{figure}

\Cref{fig:pipeline} separates raw acquisition from the first governed operation: committing the source as \code{episode.add}.  Optional claim admission and all later lifecycle changes append events to the same authenticated ledger, as detailed in \cref{tab:transitions}.  Public projection rechecks source-binding eligibility, applies conflict and current non-revival policy, and emits assertable facts plus testimony-only episodes.  Ranking only orders those candidates.  An external component proposes structured claims; the local release decision is a separate protocol rather than another retrieval score.

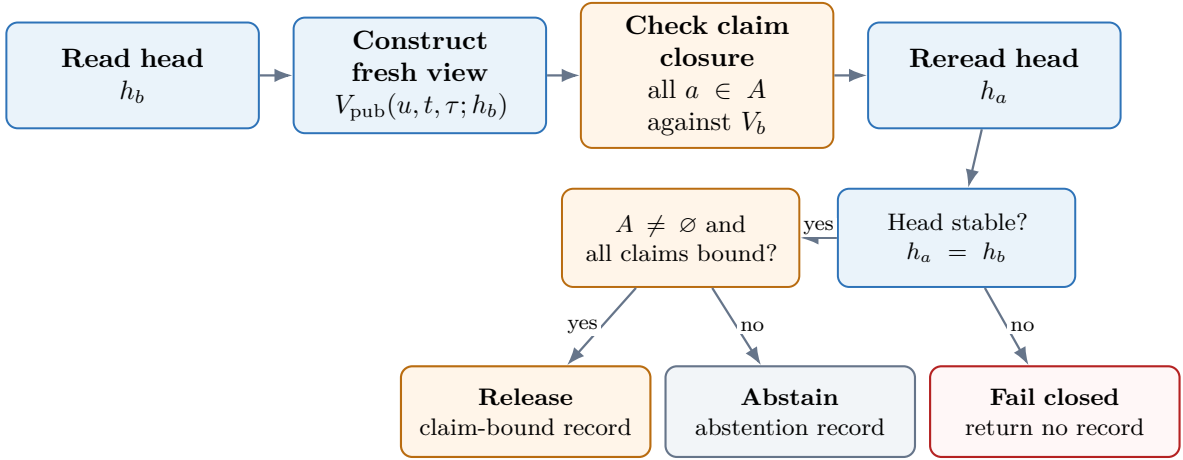
\begin{figure}[H]
\centering
\begin{tikzpicture}[
  >=Latex,
  proc/.style={draw,rounded corners=1.5mm,text width=30mm,minimum height=14mm,align=center,line width=.75pt,font=\small,inner sep=1.7mm},
  decision/.style={draw,rounded corners=1.5mm,text width=29mm,minimum height=13mm,align=center,line width=.75pt,font=\footnotesize,inner sep=1.2mm},
  outcome/.style={draw,rounded corners=1.5mm,text width=30mm,minimum height=12mm,align=center,line width=.8pt,font=\footnotesize,inner sep=1.5mm},
  flow/.style={->,line width=.85pt,draw=rankgray},
  edge/.style={font=\scriptsize,fill=white,inner sep=1pt}
]
  \node[proc,draw=govblue,fill=govfill] (hb) at (-5.7,1.7) {\textbf{Read head}\\$h_b$};
  \node[proc,draw=govblue,fill=govfill] (view) at (-1.9,1.7) {\textbf{Construct fresh view}\\$V_{\mathrm{pub}}(u,t,\tau;h_b)$};
  \node[proc,draw=releaseorange,fill=releasefill] (closure) at (1.9,1.7) {\textbf{Check claim closure}\\all $a\in A$ against $V_b$};
  \node[proc,draw=govblue,fill=govfill] (ha) at (5.7,1.7) {\textbf{Reread head}\\$h_a$};

  \node[decision,draw=govblue,fill=govfill] (stable) at (5.2,-.45) {Head stable?\\$h_a=h_b$};
  \node[decision,draw=releaseorange,fill=releasefill] (closed) at (1.55,-.45) {$A\neq\varnothing$ and\\all claims bound?};
  \node[outcome,draw=releaseorange,fill=releasefill] (released) at (-.5,-2.75) {\textbf{Release}\\claim-bound record};
  \node[outcome,draw=rankgray,fill=rankfill] (abstain) at (3.0,-2.75) {\textbf{Abstain}\\abstention record};
  \node[outcome,draw=boundaryred,fill=red!3] (fail) at (6.5,-2.75) {\textbf{Fail closed}\\return no record};

  \draw[flow] (hb) -- (view);
  \draw[flow] (view) -- (closure);
  \draw[flow] (closure) -- (ha);
  \draw[flow] (ha) -- (stable);
  \draw[flow] (stable) -- node[edge,above] {yes} (closed);
  \draw[flow] (stable) -- node[edge,right] {no} (fail);
  \draw[flow] (closed) -- node[edge,left] {yes} (released);
  \draw[flow] (closed) -- node[edge,right] {no} (abstain);
\end{tikzpicture}
\caption{Fail-closed structured-release protocol.  Closure is evaluated over the view bound to $h_b$, and the local record commits to $D_A$, policy and normalizer versions, query context, snapshots, and head.  Equality of the independently reread $h_a$ is a precondition for either release or abstention; a mismatch returns no record.}
\label{fig:release-protocol}
\end{figure}

\Cref{fig:release-protocol} makes the head-stability test explicit.  The gate reads $h_b$ before constructing the public view, checks the complete structured claim set against that view, binds its digest $D_A$, and rereads $h_a$ before returning a local decision record.  A head race is an error, not an abstention; with a stable head, an empty or nonclosed set abstains.

\subsection{Strict, incremental, and segmented engines}

The strict single-file baseline rereads and verifies the full chain before each append batch and public read.  The incremental engine verifies the full prefix at load, validates only new events against the current head on append, and checks the trusted head plus disk state on public reads.  A full audit still rereads all events and validates arbitrary-position tampering, sequence, hashes, provenance, and supersession.  With the same trusted time input, the two engines produce byte-identical logical ledgers.

The segmented engine uses fixed event-capacity segments, a serialized single-writer lock, transaction intents and receipts, and startup recovery for unfinished writes.  User-index shards contain event references.  The manifest authenticates each shard by record count and rolling hash; target-shard verification precedes user projection, and full audit covers the union of declared and physically present shards.  Recovery rebuilds both index content and commitments from authoritative segments.  The design is local concurrency control, not distributed consensus.

\section{Experimental Method}

\subsection{Research questions and evidence hierarchy}

We ask five questions.  First, does the complete I1--I5 contract distinguish matched controls from targeted violations under evaluator-hidden execution?  Second, are the clauses mutually necessary within the benchmark's constructed contract surface?  Third, do a production-code-independent finite reference model and three production implementations agree within stated bounds?  Fourth, does the contract hold end to end through the real ingestion and release services on its explicit command and query surface, including against an ungoverned local 7B comparison?  Fifth, does the governance layer preserve utility when a public task contains none of the governed interventions?  Evidence is ordered accordingly: frozen hidden contract tests and sealed end-to-end service tests are primary bounded empirical evidence; bounded exploration and implementation differential are finite formal and engineering evidence; performance is descriptive; clean public tasks are utility controls; and model-judged answer labels are admitted only if an internally prespecified reliability gate passes.

\subsection{GPM-ReleaseBench v1}

GPM-ReleaseBench v1 is an internally designed contract-conformance suite with 12 categories: source misbinding, cross-user mixing, scalar conflict, delayed-arrival bitemporality, retraction non-revival, deletion non-revival, deletion-index integrity, illegal overwrite, release-head change, exact claim closure, valid multivalue state, and unsupported free-text extension.  The public development split contains 360 cases.  The evaluator-hidden split contains 3,600 cases, 300 per category, arranged as 1,800 strict counterfactual pairs.  Pair members differ by the smallest constructed change intended to move one behavior across a contract boundary.  Runtime inputs contain opaque case identifiers and operations only; category, polarity, pair identifier, criticality, applicable-system metadata, and acceptable outcomes are withheld from runners.

The generator, task manifest, metric definitions, system runners, and stopping rules were frozen before hidden execution.  Two isolated AI review lines independently audited the hidden gold and returned GO; no human blind annotation was performed.  Hidden gold was then sealed with AES-256-GCM, its plaintext evaluator file removed, and the runner executed without the evaluator key.  This is process and cryptographic separation on one local host, not separate-user or hardware isolation.  The benchmark is evaluator-hidden relative to the tested runners, but it was designed by the same project as \sys{} and is not an external naturalistic distribution.

\subsection{Matched systems and deterministic outcomes}

We run four complete systems.  \emph{Raw append} exposes the accumulated stream.  \emph{Latest-first} resolves a key by its most recent candidate.  \emph{Flat conflict-preserving} blocks direct unresolved conflicts but lacks the full provenance, bitemporal, index-integrity, and stable-head release contract.  \sys{} implements I1--I5.  A case is an \emph{atomic match} only if its complete structured output matches one allowed outcome bundle.  \emph{Answer-bundle accuracy} excludes violation cases whose only purpose is to diagnose unsupported free text.  \emph{Unsafe release} is an unmatched \textsc{release} on a violation-polarity case.  \emph{Safe coverage} is the fraction of cases with at least one acceptable release bundle for which the system returns such a release; choosing an allowed fail-closed outcome does not count as coverage.

The internally prespecified comparison first considers complete valid policies with zero unsafe release in every critical category and, if any exist, maximizes safe coverage, breaking ties by answer-bundle accuracy and stable identifier.  If none satisfies the zero-critical condition, it minimizes critical-category unsafe releases before applying the same safe-coverage, accuracy, and stable-identifier order.  Because the suite is a fully enumerated generated artifact with no external sampling frame, we report deterministic counts and exact within-suite point differences only.  We make no confidence-interval or population-incidence claim.

StateFuse is additionally run through its official state projection and the same structured closure adapter on the 2,100 cases supported by their shared contract surface \citep{volkov2026statefuse}.  Unsupported bitemporal, user-deletion, deletion-index, stable-head, and free-text-release cases are marked unsupported rather than scored as failures.  This shared-subset result is not eligible for the complete-system ranking.

\subsection{Sealed end-to-end governed-memory evaluation}

Governed-QA Sealed tests a complementary implementation surface.  Each arm contains 2,400 deterministically generated clusters: 300 each for direct recall, set and count, temporal recall, update and conflict, multi-hop joins, false-memory pressure, open key--value notes, and boundary pressure.  A cluster traverses the production ingestion and release services rather than a paper-only adapter.  Boundary clusters require narrative statements not to be ingested and ordinary non-memory questions to be returned to the general provider.  Governed answers are exact deterministic service outputs; no language model generates them at query time.

The candidate file set is hash-frozen before a random seed is committed and revealed.  Generation fixes each machine oracle at case construction.  Answerable cases require byte-exact answers, unanswerable cases require abstention, and ordinary questions require pass-through; wrong answers, abstention on answerable cases, unsafe release, and swallowed ordinary questions are distinct failures.  Refusing everything therefore cannot score well.  The V5 freeze additionally closes the statically reachable implementation set, pins the generation date, and binds 44 files spanning ingestion, release, grammar, lineage, policy, and scoring.

The comparison receives the same setup turns and questions but answers them directly with local Qwen2.5-7B through Ollama at temperature zero, without governed storage or release.  Its scorer is deliberately generous: an answerable response is correct when it contains every bare gold value, Chinese numerals count for count questions, and ordinary questions are always credited.  The same local 7B may nominate candidate spans for optional free-form capture, but deterministic validation decides admission and the governed query answer remains model-independent.

We retain the original V3 arm as the historical public disclosure \citep{aethmere2026governedqa}.  V3 crossed UTC dates between candidate and baseline execution, producing 120 temporal questions whose date literal differed although slice and generated gold were identical.  Its reducer was therefore amended after freeze to admit only that dated form and records the amendment in its output.  V5 is a separate methodological confirmation: it pins the date before execution, reports zero question-surface mismatches and \code{reducer\_amended\_after\_freeze=false}, and evaluates Chinese- and English-command arms separately.  One-sided 95\% Clopper--Pearson lower bounds summarize the observed all-success denominators; they do not turn this internally generated contract surface into a natural-population sampling frame.

\subsection{Finite verification and implementation differential}

A separate executable reference model imports no production engine code.  Within explicit finite domains it enumerates two users, two entities, two keys, three normalized value representatives, three valid-time points, two facts, ledgers of up to four events, four single-fault classes, and five commit crash phases.  The full contract and six guard-removal mutants---source binding, valid time, conflict, lifecycle barrier, source closure, and stable head---are checked at every query state.  This is bounded exhaustive model checking, not a TLA+/Alloy result or an unbounded proof.

A separate deterministic random-schedule test compares the strict in-memory production projection, incremental in-memory production engine, and segmented persistent production engine.  Each of 100,000 traces uses a unique user to prevent cross-trace semantic interference.  After one complete audit, the segmented engine uses a frozen in-memory user-index adapter over the unchanged production projection to make all reads tractable.  Head races and physical crash schedules remain in targeted suites and the bounded model rather than this bulk test.

\subsection{Performance design}

Measurements run on Windows x64 with Node.js v24.18.0, an Intel Core i9-13900H (20 logical CPUs), and 63.64 GiB RAM.  Six paired fresh-process repetitions run strict and incremental engines in balanced order on the same 10,008-event mixed workload, using 2,000-event append batches.  Six segmented repetitions ingest one million synthetic episode events for 1,000 users, warm 20 searches, time 200 searches, and execute audit, cold start, tamper detection, and recovery.  Medians and full ranges are descriptive; six machine runs are not IID population samples.

\subsection{Public clean-task controls}

We use LongMemEval and MemoryAgentBench EventQA only as clean-task controls \citep{wu2025longmemeval,hu2025memoryagentbench}.  EventQA rows 0--9 were inspected during integration and excluded; rows 10--16 remained eligible.  LongMemEval cases were selected from a 347-item project-fresh lock.  Project-fresh means not used in this project's prior development or tuning; public benchmark exposure during model pretraining is not claimed absent.

Version 1 freezes 120 LongMemEval and 120 EventQA cases before outcomes and uses Qwen2.5-1.5B and Llama-3.2-1B.  After its semantic-judge gate fails, version 2 is opened as a separately versioned confirmation on 240 disjoint cases and uses Qwen2.5-7B and Gemma-3-4B.  EventQA is balanced between 65,536- and 131,072-token strata.  Both versions use temperature zero, top-12 retrieval, and a 24,000-character evidence budget.  On every selected clean task, flat-conflict and \sys{} produce content-hash-identical candidate pools, so one shared generator call represents both conditions.  Their answer difference is therefore structural, not an independently estimated model effect.

EventQA uses the upstream deterministic substring criterion.  LongMemEval uses two local AI judges with generator identity hidden.  The internally prespecified primary gate requires Cohen's $\kappa\geq0.80$ separately for correctness, hallucination, and omission.  Failure leaves all LongMemEval answer-layer rates descriptive.

\section{Results}

\subsection{Evaluator-hidden contract results}

\begin{table}[H]
\centering
\small
\caption{GPM-ReleaseBench v1 evaluator-hidden results on 3,600 cases.  Unsafe release is conditioned on 1,800 violation-polarity cases; other denominators follow the definitions in the text.}
\label{tab:hidden}
\begin{tabularx}{\textwidth}{@{}Yrrrr@{}}
\toprule
System & Atomic match & Answer acc. & Unsafe release & Safe coverage \\
\midrule
Raw append & 1500/3600 & 43.48\% & 91.67\% & 76.92\% \\
Latest-first & 1500/3600 & 43.48\% & 58.33\% & 53.85\% \\
Flat conflict-preserving & 1800/3600 & 52.17\% & 50.00\% & 61.54\% \\
Full \sys{} & 3600/3600 & 100.00\% & 0.00\% & 84.62\% \\
\bottomrule
\end{tabularx}
\end{table}

Flat conflict-preserving is the strongest of the three intentionally simple complete policies under the frozen rule.  Relative to it, \sys{} changes unsafe release by $-50.00$ percentage points, answer-bundle accuracy by $+47.83$ points, safe coverage by $+23.08$ points, and atomic match by $+50.00$ points.  The difference exceeds the internally predefined deterministic margin of $-2$ answer-accuracy points.  These are deterministic differences for the generated hidden suite, not estimates of natural-world failure frequency.  Each of five critical violation categories has 0/150 unsafe releases; this is an exact suite count, not a population confidence statement.

On StateFuse's 2,100 supported shared-contract cases, the official projection plus common closure adapter matches 1,200 complete bundles, has no unmatched violation release, and has 62.50\% safe coverage.  Because 1,500 cases are outside its declared adapter surface, this does not establish an overall ranking against \sys{}.

\subsection{End-to-end governed-memory results}

\begin{table}[H]
\centering
\small
\caption{Governed-QA Sealed V3 end-to-end results.  The ungoverned comparison is local Qwen2.5-7B at temperature zero; the governed lane traverses the deterministic ingestion and release services.}
\label{tab:sealed-v3}
\begin{tabularx}{\textwidth}{@{}Yrr@{}}
\toprule
Query family & Ungoverned 7B & Governed lane \\
\midrule
Direct recall & 41/300 (13.7\%) & 300/300 \\
Set and count & 98/300 (32.7\%) & 300/300 \\
Temporal recall & 63/300 (21.0\%) & 300/300 \\
Update and conflict & 41/300 (13.7\%) & 300/300 \\
Multi-hop joins & 65/300 (21.7\%) & 300/300 \\
False-memory pressure & 45/300 (15.0\%) & 300/300 \\
Open key--value notes & 34/300 (11.3\%) & 300/300 \\
Boundary pressure$^{\ast}$ & 213/300 (71.0\%) & 300/300 \\
\midrule
Total & 600/2,400 (25.0\%) & 2,400/2,400 (100\%) \\
\bottomrule
\end{tabularx}
\vspace{2pt}
\raggedright\footnotesize $^{\ast}$Ordinary non-memory questions are always credited to the ungoverned baseline because it is supposed to answer them; this makes its boundary-family score comparatively high.
\end{table}

The governed lane is correct on all 2,400 clusters, with a one-sided 95\% Clopper--Pearson lower bound of 99.875\%.  Of the 1,800 clusters missed by the ungoverned 7B, all 1,800 are corrected; none of its 600 correct clusters regresses.  The bounded cure lower bound is 99.834\%.  These counts do not mean that Qwen2.5-7B reached 100\% accuracy.  The model is the 25\% ungoverned comparison; the 100\% result belongs to a deterministic governed lane on its explicit contract.

\begin{table}[H]
\centering
\small
\caption{Governed-QA Sealed V5 confirmation after generation-date pinning.  ``Repaired'' is conditioned on baseline failure; regressions are conditioned on baseline-correct clusters.}
\label{tab:sealed-v5}
\begin{tabularx}{\textwidth}{@{}Yrrrr@{}}
\toprule
Command arm & Ungoverned correct & Governed correct & Repaired & Regressions \\
\midrule
Chinese & 1,346/2,400 (56.1\%) & 2,400/2,400 & 1,054/1,054 & 0/1,346 \\
English & 636/2,400 (26.5\%) & 2,400/2,400 & 1,764/1,764 & 0/636 \\
\bottomrule
\end{tabularx}
\end{table}

Both V5 command arms again pass every 300-case family gate.  The correctness lower bound is 99.875\% per arm; cure lower bounds are 99.716\% for Chinese commands and 99.830\% for English commands.  Unlike V3, V5 requires no post-freeze reducer amendment and has no date-surface mismatch.  At preparation of this revision, all 44 frozen candidate files, the reachable-file closure, and the seed commitment reverified without drift.

\subsection{Post-aggregate guard-removal diagnostics}

The single-clause removals in \cref{tab:exploratory} were implemented only after aggregate full-system and policy results were visible.  They are mechanism diagnostics, not prespecified hidden ablations or unbiased causal estimates.

\begin{table}[H]
\centering
\small
\caption{Post-aggregate exploratory removal of one obligation.}
\label{tab:exploratory}
\begin{tabular}{@{}lrrl@{}}
\toprule
Condition & Atomic match & Unsafe releases & Exposed contract surface \\
\midrule
No I1 & 3450/3600 & 0 & index commitment detection \\
No I2 & 3450/3600 & 150 & source-binding/supersession admission \\
No I3 & 3450/3600 & 150 & unresolved scalar conflict \\
No I4 & 3000/3600 & 450 & retraction and deletion barriers \\
No I5 & 2850/3600 & 750 & closure, fresh view, stable head \\
\bottomrule
\end{tabular}
\end{table}

No-I1 illustrates why unsafe-release count alone is insufficient: the mutant misses 150 required fail-closed outcomes without directly releasing an unmatched claim.  Every contract clause is therefore evaluated through complete outcome bundles rather than one safety rate.

\subsection{Finite verification and three-engine agreement}

\begin{table}[H]
\centering
\small
\caption{Finite executable verification.  Counts are exhaustive or deterministic within the stated bounds.}
\label{tab:finite}
\begin{tabularx}{\textwidth}{@{}Y r@{}}
\toprule
Check & Result \\
\midrule
Semantic states / query states & 331,776 / 1,990,656 \\
Full-contract counterexamples & 0 \\
Guard removals with a counterexample & 6/6 \\
Valid finite ledgers / injected faults rejected & 30 / 121 \\
Differential traces / logical actions & 100,000 / 287,395 \\
Engine action executions / observation points & 862,185 / 300,000 \\
Segmented persisted events / full-audit result & 349,988 / pass \\
Public, audit, or release mismatches & 0 \\
\bottomrule
\end{tabularx}
\end{table}

The independent model finds no counterexample to the full contract and finds at least one for every guard removal.  Consistent with the valid-ledger definition above, it rejects a decreasing transaction time, as well as tamper, truncation, reordering, duplication, and invalid crash-recovery states, within its bounds.  The production differential in \cref{tab:finite} observes no mismatch.  Zero bounded counterexamples and zero sampled implementation differences remain finite evidence, not a proof that the implementation is defect-free.

All 39 incremental, 41 segmented, 24 retraction-barrier, 39 trust-boundary, and 22 release-contract implementation checks also pass.  The added release checks reject claim or source-identifier substitution and stale-head replay; applying the strengthened record schema leaves all 3,600 hidden benchmark decisions byte-identical.  Four writer processes complete 120 events without observed loss, duplication, or fork; this is only a lock smoke test.

\subsection{Performance and scale}

\begin{table}[H]
\centering
\small
\caption{Six fresh-process repetitions.  Values are descriptive median and full range.}
\label{tab:performance}
\begin{tabular}{@{}lrr@{}}
\toprule
Metric & Median & Range \\
\midrule
Strict single-file throughput (events/s) & 12,295.2 & 10,506.3--12,583.7 \\
Incremental throughput (events/s) & 44,156.4 & 34,580.0--49,571.3 \\
Paired speedup & $3.565\times$ & $2.809$--$4.718\times$ \\
Strict-first paired speedup & $3.725\times$ & $3.688$--$4.718\times$ \\
Incremental-first paired speedup & $3.303\times$ & $2.809$--$3.442\times$ \\
Segmented 1M throughput (events/s) & 39,300.8 & 38,352.9--40,099.6 \\
Segmented full audit (s) & 8.702 & 8.538--9.527 \\
Cold-start full audit (s) & 8.638 & 8.526--9.495 \\
Search p50 / p95 (ms) & 57.332 / 70.537 & 57.046--61.702 / 65.027--78.801 \\
Peak RSS (MiB) & 340.3 & 336.5--346.1 \\
\bottomrule
\end{tabular}
\end{table}

All six single-file pairs produce identical logical-ledger hashes, and every segmented run passes full verification, tamper detection, and recovery.  The speedup in \cref{tab:performance} comes from avoiding repeated audit of an already verified prefix, not from deleting the audit path.  Nonoverlapping order strata indicate a systematic cache or warm-up effect; both strata and the pooled median are therefore shown.  Segmented episode-only stress and mixed single-file workloads are not an absolute architecture comparison.

\subsection{Clean public tasks and failed language gate}

\begin{table}[H]
\centering
\small
\caption{Deterministically scored EventQA clean-task controls.  ``Substring match'' is the upstream criterion, not semantic exact match.  Each row is one shared generation because flat-conflict and \sys{} prompts are byte-identical.}
\label{tab:eventqa}
\begin{tabular}{@{}llrr@{}}
\toprule
Version & Local generator & Substring match & Abstention \\
\midrule
V1 & Qwen2.5-1.5B & 41/120 (34.17\%) & 4/120 (3.33\%) \\
V1 & Llama-3.2-1B & 3/120 (2.50\%) & 2/120 (1.67\%) \\
V2 & Qwen2.5-7B & 69/120 (57.50\%) & 25/120 (20.83\%) \\
V2 & Gemma-3-4B & 59/120 (49.17\%) & 0/120 (0.00\%) \\
\bottomrule
\end{tabular}
\end{table}

\begin{table}[H]
\centering
\small
\caption{LongMemEval semantic-judge reliability.  Primary eligibility required every label-specific kappa to reach 0.80.}
\label{tab:judgegate}
\begin{tabular}{@{}lrrrrc@{}}
\toprule
Version & Parse valid & Full agreement & $\kappa_{\rm correct}$ & $\kappa_{\rm hall.}$ / $\kappa_{\rm omit}$ & Eligible \\
\midrule
V1 & 97.50\% & 10.26\% & .0734 & $-.0053$ / .0282 & no \\
V2 & 100.00\% & 75.83\% & .8090 & .1705 / .7567 & no \\
\bottomrule
\end{tabular}
\end{table}

The structural flat-versus-\sys{} answer difference is 0 points in both versions and exceeds the internally predefined deterministic margin of $-2$ points, but it is not a stochastic treatment estimate.  \Cref{tab:eventqa} measures local-model utility only and makes no frontier-model claim.  The LongMemEval answer counts---V1: 32/120 and 27/120; V2: 42/120 and 18/120 for the two respective generators---are retained as descriptive negative evidence.  Because \cref{tab:judgegate} fails in both versions, no LongMemEval answer-accuracy, hallucination, or omission rate supports a primary claim.

The candidate reproduction artifact contains the public development benchmark, finite checks, frozen locks, and aggregate public-task receipts, but excludes hidden gold, keys, licensed raw datasets, model weights, and raw prompts.  Two independently copied same-host fresh-directory runs reproduce the 360-case development scores, bounded counts, and 100,000-trace differential receipt with the same semantic SHA-256.  This verifies packaging determinism on one host; it is not independent third-party reproduction.

\section{Related Work and Discussion}

Retrieval-augmented generation and long-term agent memory primarily optimize access to information beyond the current context \citep{lewis2020rag,packer2023memgpt,wang2023longmem,chhikara2025mem0,li2025memos}.  Recent systems add temporal graphs, consolidation, learned memory operations, and reflective structure \citep{banerjee2026apex,hu2026evermemos}.  Text2Mem defines typed executable operations across memory backends \citep{wang2026text2mem}, while conflict-aware memory applies explicit detection rules to vector data quality \citep{ma2026conflict}.  Long-memory benchmarks now cover dialogue recall, updating, forgetting, hallucination propagation, and controlled conflict regimes \citep{maharana2024locomo,wu2025longmemeval,chen2025halumem,hu2025memoryagentbench,tao2026memconflict}.  A controlled ACL study further shows that addition, deletion, error propagation, and experience replay policies change agent behavior in distinct ways \citep{xiong2026memorymanagement}.

Adjacent formal work separates three decisions that ordinary retrieval leaves implicit.  Trust-sensitive revision filters a report through source- and domain-specific trust before changing belief \citep{booth2018trust}; revision by history constrains current revision through a sequence of prior revisions and their outcomes \citep{liberatore2015history}; and partial abstention optimizes the choice to withhold uncertain components of a multilabel prediction \citep{nguyen2021partial}.  \sys{} does not replace these belief-revision or decision-theoretic accounts.  It gives an executable systems contract for a different boundary: whether source-bound, bitemporal, lifecycle-governed state may support an explicit outgoing claim under a fresh verified ledger head.

A close contract-level neighbor is StateFuse, which builds deterministic conflict objects, projection-scoped resolution, and exact or semantic retraction handles over a replicated OpSet/CRDT substrate \citep{volkov2026statefuse}.  Its reported evidence shows that conflict-preserving surfaces can expose disagreement without a general answer-accuracy advantage over strong flat baselines.  \sys{} is not replica-convergent and has no unseen-target semantic tombstone.  In return, its evaluated contract starts earlier, by binding admitted claims to source content and valid/transaction time, and ends later, by requiring a fresh public projection and verified head before an explicit claim set can be released.

MemoRepair studies a complementary lifecycle stage: deletion, correction, or interface migration can invalidate arbitrary summaries, caches, skills, and tool procedures downstream of a source \citep{zhao2026memorepair}.  Its influence-provenance graph, withdrawal barrier, validated successor construction, and predecessor-closed repair selection are outside \sys{}'s scope.  I4 covers the target fact, its bound episode, and engine-declared indexes and projections; I4 therefore does not provide cascade repair for arbitrary derived artifacts.  Conversely, MemoRepair does not specify the bitemporal claim-admission and structured answer-release closure studied here.

TOKI and MemIR are adjacent at different layers.  TOKI supplies explicit bitemporal operators for contradiction-aware state reconstruction and resolution \citep{wang2026toki}; \sys{} instead fixes one public-policy interpretation and tests its non-revival behavior in three engines.  MemIR addresses provenance-role collapse by keeping evidence, assertion, and decision objects typed \citep{jin2026memir}; \sys{} binds source evidence at admission but does not offer MemIR's richer role system.  Neither distinction implies empirical superiority because the released tasks and contracts differ.

\begin{table}[H]
\centering
\small
\caption{Contract-level positioning of closely related systems.  Cells summarize the contract each work makes explicit; they are not an empirical ranking.}
\label{tab:related-contracts}
\begin{tabularx}{\textwidth}{@{}P{.13\textwidth}P{.24\textwidth}P{.28\textwidth}Y@{}}
\toprule
System & Governed state & Change or invalidity contract & Outgoing boundary \\
\midrule
StateFuse \citep{volkov2026statefuse} & Replicated immutable operation set and projection & Explicit conflict objects; exact and semantic correction handles & Projection selects a candidate or abstains \\
\addlinespace
TOKI \citep{wang2026toki} & Bitemporal persistent-memory history & Operator algebra reconstructs and resolves contradictory states & Returns operator-defined state; no claim-bound local decision \\
\addlinespace
MemIR \citep{jin2026memir} & Typed evidence, assertion, and decision records & Role constraints prevent provenance from collapsing into unsupported state & Preserves typed lineage; no stable-head local release record \\
\addlinespace
MemoRepair \citep{zhao2026memorepair} & Provenance-linked derived artifacts & Barrier-first withdrawal and validated descendant repair & Republishes validated, predecessor-closed successors \\
\addlinespace
\sys{} & Source-bound bitemporal claims and episodes & Current barriers, conflict isolation, supersession, and retraction & Claim-bound local record over a fresh view and stable head \\
\bottomrule
\end{tabularx}
\end{table}

Provenance graphs, bitemporal databases, conflict detection, and selective prediction are established components with mature precedents \citep{w3c2013prov,snodgrass1999time,geifman2017selective}.  The ledger and head-reread design likewise sits beside established work on hash-linked timestamping, append-only transparency logs, optimistic validation, and mutable usage-control decisions \citep{haber1991timestamp,laurie2021ct,kung1981optimistic,park2004ucon}.  \sys{} does not replace those security or database mechanisms.  Its contribution is their scoped composition into one public-eligibility decision: failed source binding, invalid supersession, unresolved conflict, current retraction/deletion, or a nonclosed structured claim causes the public path to fail closed.  This governance layer can be combined with different retrievers, memory graphs, operation languages, replicated conflict substrates, or descendant-repair planners.

Append-only audit and deletion rights remain in tension.  Physical event removal destroys audit evidence, whereas indefinite plaintext retention violates data minimization and user expectations.  Our result is a public-view barrier plus a separately routed audit view.  It is not media erasure, backup deletion, cryptographic erasure, or machine unlearning.  Production deployments still require retention policy, least-privilege audit authorization, encryption and key destruction where appropriate, and legally compliant deletion procedures.

Selective prediction frames coverage jointly with the risk among released outputs \citep{geifman2017selective}.  Our guard-removal and answer diagnostics show both sides: weakening gates can expose unsafe release, while excessive abstention converts errors into omissions.  The objective is not maximum abstention but maximum valid coverage under an explicit risk budget, with distinctions among conflict explanation, clarification, hedged response, and complete abstention.

\section{Limitations and Threats to Validity}

\paragraph{External validity and benchmark construction.} GPM-ReleaseBench is synthetic and was designed by the same project as \sys{}.  Its hidden split tests runner behavior against a frozen contract, not transfer to natural dialogue, new ontologies, adversarial authors, or multi-year privacy drift.  Counterfactual homogeneity makes the within-suite differences exact but says nothing about the frequency of those failures outside the suite.  The one-million-event workload is likewise engineering stress, not a natural event distribution.

\paragraph{Baselines.} The three complete baselines are intentionally simple matched policies, not full Mem0, MemGPT, MemOS, TOKI, MemIR, MemoRepair, or graph-memory systems.  StateFuse is evaluated only on a 2,100-case shared-contract subset.  Different systems expose different ingestion, invalidation, replication, and release contracts, so \cref{tab:hidden} supports superiority only over the listed complete policies on this constructed suite.  The related-work matrix in \cref{tab:related-contracts} is a scope comparison, not an empirical ranking.

\paragraph{Hidden-evaluation process.} Two isolated AI review lines audited the hidden gold, but there was no independent human annotation.  Gold sealing and key separation occurred under one operating-system user, so they reduce accidental access and bind artifacts without constituting an independent security principal.  The system implementation and benchmark ontology were both available before the hidden run; evaluator-hidden execution is not the same as an externally authored benchmark.

\paragraph{Sealed service evaluation.} Governed-QA Sealed is also internally generated and tests an explicit command grammar and eight query families, not arbitrary dialogue.  Its deterministic oracle is strong for that surface but does not assess semantic equivalence outside the generated forms.  Qwen2.5-7B is one deliberately ungoverned local comparison, not the strongest possible memory system or a representative sample of model providers.  The V3 reducer amendment for 120 cross-date temporal surfaces was made after freeze and is reported rather than hidden; V5 removes that defect by pinning the date before execution.  The Clopper--Pearson bounds summarize all-success denominators within these prespecified runs and do not justify an open-world or product-wide accuracy claim.

\paragraph{Derived-artifact scope.} I4 covers facts, their bound episodes, and the indexes and materialized projections declared by the three engines.  It does not discover or repair arbitrary downstream summaries, cached outputs, embeddings, skills, or model parameters.  Such descendants require complete influence provenance and a cascade-withdrawal and repair contract of the kind studied by MemoRepair.

\paragraph{Schema coverage.} Unregistered attributes are treated as single-valued and only a small registry is multi-valued.  Normalization is NFKC, trimming, and case folding.  Set merge, vague time, unit conversion, entity coreference, and open-text contradiction require stronger schema and semantic layers.

\paragraph{Temporal policy.} Current public release uses \code{transactionAt=null}.  Non-null \code{transactionAt} is an explicit policy-filtered historical mode: it combines a requested transaction prefix with retraction and deletion policy at the current verified head.  This prevents revival of retracted or deleted state, but conflicts and valid superseding assertions written after the requested prefix are not projected into that prefix.  Consequently caller-selected historical mode may expose a fact later contradicted or superseded and is unsuitable for current-state answers.  The query hash does not authorize historical mode; deployment policy must restrict it.  Ordinary as-of reconstruction is confined to the separately authorized audit interface.

\paragraph{Release scope.} I5 decides exact matches for explicit structured claims and supplied source fact identifiers in the selected public mode.  The local record binds the canonical complete claim multiset, versions, context, snapshots, and head, and verification rejects claim substitution or a stale head.  It is not a signed cross-boundary credential.  The query hash does not prove query--claim relevance, I2 does not prove semantic entailment from quoted text, and the gate cannot detect implicit assertions in final prose.

\paragraph{Threat model.} The trusted computing base comprises engine code, system clock, single-writer coordination, deployment authentication, and a head commitment or segmented manifest protected separately from authoritative storage.  Public reads detect ordinary file replacement, equal-length modification, index-reference deletion or injection, and truncation.  The prototype does not resist a malicious authorized writer, a host attacker who rewrites both storage and commitments, forged filesystem metadata, or lack of an external anchor.  Separate audit methods are not access control, and local multi-process tests are not distributed consensus.

\paragraph{Finite verification.} The reference model is exhaustive only within its declared small domains and event bound.  The five obligations are executable interface clauses, not an inductive proof over all reachable unbounded states.  The 100,000-trace differential uses unique users and a fixed public seed; it does not cover unbounded traces, distributed interleavings, hash collisions, or a host that rewrites both storage and its independent commitment.  The post-aggregate I1--I5 removals are exploratory and cannot be treated as prespecified hidden ablations.

\paragraph{Public answer controls.} Flat-conflict and \sys{} receive byte-identical clean-task prompts, so the observed zero difference is an implementation sanity check rather than an estimated treatment effect.  EventQA values use the upstream substring criterion, not semantic exact match.  Both LongMemEval versions fail the label-wise judge-agreement gate and cannot support primary answer-accuracy, hallucination, or omission claims.  Public benchmark pretraining exposure is not claimed absent, and the local 1B--7B models do not estimate frontier performance.  These clean-task controls and the governed-query service test answer different questions: the former retains free-form model generation, whereas the latter returns deterministic contract values.

\paragraph{Reproducibility.} Two fresh-directory copies reproduce the candidate package on one host.  Hidden gold, evaluator keys, licensed raw public datasets, model weights, and raw model prompts are excluded, so the package cannot independently rescore the hidden split or rerun the language-model controls.  There is no external-team reproduction.  The minimal artifact now has an explicit PolyForm Noncommercial 1.0.0 license and passes clean extraction, but it remains release-ready rather than published: no public URL, DOI, or release date has been assigned.  The V3/V5 aggregate service disclosure is available in an immutable public repository commit \citep{aethmere2026governedqa} and is mirrored in this version's ancillary JSON; complete per-cluster receipts are not included in either the arXiv archive or the v1.0.0 artifact.

\section{Ethics, Disclosure, and Artifact Statement}

Persistent memory lengthens the lifetime of both error and privacy exposure.  Systems should reject unsupported claims by default, let users inspect, correct, and delete memory, distinguish source statements from inference, and propagate deletion into public indexes.  A conflict need not imply user dishonesty; time, context, and source error can produce apparent contradiction.

No new human subjects or private conversations from project users were used.  GPM-ReleaseBench and Governed-QA Sealed are synthetic.  LongMemEval and MemoryAgentBench are represented by upstream revisions, hashes, selected identifiers, and aggregate results subject to their licenses; their raw licensed records are not included in the artifact candidate.

The author is employed by Qingdao Guodongxiansheng Network Technology Co., Ltd. (Gu\v{o}d\`ong Xi\=ansheng).  The author and company may have financial interests in Aethmere-OS, related software, commercialization, and intellectual-property applications.  Because the author writes primarily in Chinese and has limited English proficiency, OpenAI Codex assisted with English translation, language editing, and LaTeX typesetting; it also assisted with literature organization, experiment scripting, and deterministic result checks under the author's review.  Codex is not an author.  Guodong Xu conceived the study, determined the claims and interpretation, verified the reported results and citations, reviewed and approved all AI-assisted content, and accepts full responsibility for the paper.  No external grant supported this work; the author and company supplied the local hardware and compute.

The TeX compilation inputs are limited to \code{main.tex} and \code{references.bib}.  Version 2 additionally includes \code{anc/governed-qa-v3-v5.json} as ancillary material: it mirrors the public aggregate counts, claim limits, and integrity hashes, but contains no private questions, prompts, model outputs, or per-cluster receipts.  A separate manifest-bound v1.0.0 reproducibility artifact contains the public development benchmark, executable finite checks, frozen locks, and aggregate receipts, and excludes evaluator secrets and licensed raw datasets.  It is licensed under PolyForm Noncommercial 1.0.0 and passes a same-host clean-extraction rerun, but remains release-ready rather than publicly uploaded; its URL, DOI, and release date are unset.  Large per-question records, the Governed-QA Sealed per-cluster receipts, and raw licensed datasets are intentionally excluded from the arXiv archive.

\section{Conclusion}

Governed persistent memory reframes long-term agent memory as a source-bound, bitemporal state system with an explicit public release boundary.  Admission, conflict, retraction, deletion, and release each have executable preconditions and observable failure evidence.  On a frozen 3,600-case contract suite, the complete system matches every allowed outcome and avoids unmatched violation release, while the strongest of three intentionally simple complete policies matches half of the cases and releases unsafely on half of the violation cases.  On the separate sealed end-to-end service surface, the V3 governed lane returns 2,400/2,400 contract-correct outcomes versus 600/2,400 for an ungoverned local 7B and repairs all 1,800 baseline failures without regression; V5 repeats 2,400/2,400 in both command-language arms without a reducer amendment.  A production-code-independent bounded reference model and a 100,000-trace three-engine differential find no full-contract counterexample or implementation mismatch within their declared bounds.  Contract consequence~1 states the resulting source-relative non-revival property for claim-bound local records.

The same evidence sets strict limits.  Both contract benchmarks are internally designed and synthetic; finite exploration is not an unbounded proof; clean public tasks contain no governed intervention; and both LongMemEval semantic-judge protocols fail their reliability gate.  The 100\% sealed result belongs only to the explicit governed contract, not to Qwen2.5-7B, open-world dialogue, or the product as a whole.  Governed state is therefore one necessary foundation for reliable long-term memory, not a sufficient solution to relevance, truth, or free-form generation.  Externally authored challenge sets, publication of the licensed artifact and sealed receipts, independent reproduction, arbitrary descendant repair, richer typed conflict semantics, free-text evidence closure, and longitudinal user studies remain open work.

\bibliographystyle{plainnat}
\bibliography{references}

\clearpage
\appendix
\section{Claim--Evidence Boundary}

\begin{table}[h]
\centering
\small
\caption{What each main claim does and does not establish.}
\renewcommand{\arraystretch}{1.08}
\begin{tabularx}{\textwidth}{@{}P{.19\textwidth}P{.28\textwidth}P{.25\textwidth}Y@{}}
\toprule
Claim & Current evidence & Unsupported extrapolation & Falsifier \\
\midrule
I1--I5 distinguish the frozen contract suite & 3,600/3,600 complete outcome matches; 0/1,800 unmatched violation releases & Natural failure incidence, world truth, or superiority over untested full systems & A frozen case fails to match its allowed complete outcome \\
\addlinespace
The governed service satisfies its explicit command/query contract & V3: 2,400/2,400; all 1,800 baseline failures repaired, 0/600 regressions. V5: 2,400/2,400 per command arm & Open-world or product accuracy; 7B model accuracy of 100\%; superiority over governed model-based systems & A bound cluster fails, an ordinary question is swallowed, or a baseline-correct cluster regresses \\
\addlinespace
No contract counterexample is found in the declared finite model & 331,776 states, 1,990,656 query states, zero full counterexamples, six mutant counterexamples & Unbounded proof, distributed consensus, or host-compromise resistance & A full-contract counterexample appears within the declared bounds \\
\addlinespace
Three implementations agree on sampled schedules & 100,000 traces, 300,000 observations, zero public/audit/release mismatch & Absence of all implementation defects or concurrency proof & Same trace and head produce different observed semantics \\
\addlinespace
Incremental verification reduces repeated-prefix cost & Six fresh-process, balanced-order pairs with identical logical hashes & Universal speedup across hardware or workloads & Equivalent inputs diverge, or repeated trials lose the advantage \\
\addlinespace
Clean-task utility is not degraded by projection & Byte-identical flat and \sys{} prompts; deterministic EventQA controls & Safety improvement, model effect, frontier performance, or valid LongMemEval rates & Eligible clean facts differ between the two candidate pools \\
\bottomrule
\end{tabularx}
\end{table}

\end{document}